\documentclass[sigconf,10pt]{acmart}
\usepackage[utf8]{inputenc}

\usepackage{subcaption}
\usepackage{cleveref}
\usepackage{soul}
\usepackage{booktabs} 
\usepackage{tabu}
\usepackage{extarrows}
\usepackage{stmaryrd}
\usepackage{gensymb}
\usepackage{lipsum}
\usepackage{tikz}
\usepackage{xcolor}
\usepackage{float}
\usepackage{stfloats} 
\usepackage[bb=boondox]{mathalfa}
\usepackage{pifont} 

\newcommand*\circled[1]{\tikz[baseline=(char.base)]{
            \node[shape=circle,fill,inner sep=1.2pt] (char) {\textcolor{white}{#1}};}}

\newif\ifcomment

\commenttrue

\ifcomment
\definecolor{stelios_colour}{RGB}{144, 238, 144}
\newcommand{\stelios}[1]{\sethlcolor{stelios_colour}\hl{[\textbf{Stelios:} #1]}}
\definecolor{royson_colour}{RGB}{191, 232, 255}
\newcommand{\royson}[1]{\sethlcolor{royson_colour}\hl{[\textbf{Royson:} #1]}}
\else
\newcommand{\stelios}[1]{}
\newcommand{\royson}[1]{}
\fi

\usepackage{background}

\backgroundsetup{angle=0, 
scale=1,
color=black,
firstpage=true,
position=current page.north,
hshift=0pt,
vshift=-20pt,
contents={\ifnum\value{page}=1 PREPRINT: Accepted at the 1st Workshop on Distributed Machine Learning at CoNEXT 2020 (DistributedML 2020) \else \fi}
}

\AtBeginDocument{%
  \providecommand\BibTeX{{%
    \normalfont B\kern-0.5em{\scshape i\kern-0.25em b}\kern-0.8em\TeX}}}

\copyrightyear{2020}
\acmYear{2020}
\setcopyright{acmcopyright}\acmConference[DistributedML'20]{1st Workshop on Distributed Machine Learning}{December 1, 2020}{Barcelona, Spain}
\acmBooktitle{1st Workshop on Distributed Machine Learning (DistributedML'20), December 1, 2020, Barcelona, Spain}
\acmPrice{15.00}
\acmDOI{10.1145/3426745.3431336}
\acmISBN{978-1-4503-8182-6/20/12}

\begin{document}

\title{Neural Enhancement in Content Delivery Systems:\\ The State-of-the-Art and Future Directions}




\author{Royson Lee$^\dagger$*, Stylianos I. Venieris$^\dagger$*, Nicholas D. Lane$^{\dagger,\ddagger}$}
\affiliation{\institution{$^\dagger$Samsung AI Center, Cambridge\hspace{+0.75cm}$^\ddagger$University of Cambridge}{\Small\textit{{* Indicates equal contribution.}}}}

\authorwithoutinstuition{{Royson Lee, Stylianos I. Venieris, Nicholas D. Lane}}

\renewcommand{\shortauthors}{Lee and Venieris, et al.}

\begin{abstract}

Internet-enabled smartphones and ultra-wide displays are transforming a variety of visual apps spanning from on-demand movies and 360\textdegree~ videos to video-conferencing and live streaming. However, robustly delivering visual content under fluctuating networking conditions on devices of diverse capabilities remains an open problem. In recent years, advances in the field of deep learning on tasks such as super-resolution and image enhancement have led to unprecedented performance in generating high-quality images from low-quality ones, a process we refer to as neural \mbox{enhancement}. 
In this paper, we survey state-of-the-art content delivery systems that employ neural enhancement as a key component in achieving both fast response time and high visual quality. We first present the deployment challenges of neural enhancement models. We then cover systems targeting diverse use-cases and analyze their design decisions in overcoming technical challenges. Moreover, we present promising directions based on the latest insights from deep learning research to further boost the quality of experience of these systems.


\end{abstract}




\fancyhead{}
\maketitle

\section{Introduction}

Internet content delivery has seen a tremendous growth over the past few years. Specifically, video traffic is estimated to account for 82\% of global Internet traffic by 2022 – up from 75\% in 2017~\cite{ciscovni}. This growth is attributed to not only the rapid increase of Internet-enabled devices, but also the support for higher-resolution content. For instance, an estimated 66\% of TV sets will support Ultra-High-Definition (4K) videos by 2023 as compared to 33\% in 2018~\cite{ciscoreport}. Most importantly, content traffic such as live streaming, video conferencing, video surveillance, and both short- and long-form videos-on-demand, are expected to rise very quickly. 
To meet these demands, a new class of 
distributed systems has emerged. 
Such systems span from video analytics frameworks that co-optimize latency and accuracy~\cite{Wang2019, Du2020} , to content delivery systems~\cite{Mao2017,Yeo2018} that aim to maximize the quality of experience (QoE), estimated based on the chosen bitrate and amount of rebuffering.


One of the primary challenges of distributed systems for content delivery is their reliance on networking conditions. Currently, the quality of the communication channel between client and server plays a key role in meeting the application-level performance needs due to the significant amount of transferred data and the tight latency targets. Nevertheless, in real-life mobile networks, the communication speed fluctuates and poor network conditions lead to excessive response times, dropped frames or video stalling, that rapidly degrade the user experience. This phenomenon is further amplified by the increasing number of users which compete for the same pool of network resources.




One recent key method that enables tackling this challenge is \textit{neural enhancement} through super-resolution (SR) and image enhancement models. These models are capable of processing a low-resolution/quality image and generating a high-quality output. With the unprecedented performance of convolutional neural networks (CNNs), content delivery systems have begun integrating neural enhancement models as a core component. The primary paradigm of using neural enhancement models in content delivery systems comprises the transmission of compact low-resolution/quality content, often along with the associated model, followed by its subsequent quality enhancement on the receiver side through a enhance-capable model~\cite{Yeo2017}. In this manner, the transfer load is minimized, drastically reducing the network footprint and the corresponding bandwidth requirements. 

Despite their benefits, integrating state-of-the-art neural enhancement models into visual content delivery systems introduces significant challenges. First, these models, especially SR models, have excessive computational demands that are measured up to the hundreds of TFLOPs per frame in order to achieve an upscaling of up to 4K/8K.
With client platforms typically comprising devices with strict resource and battery constraints~\cite{embench_2019}, clients are still struggling to execute neural enhancement models on-device while meeting the target quality~\cite{Lee2019}. This fact is aggravated by the stringent latency and throughput requirements that are imposed in order to sustain high QoE. Finally, enabling the deployment of such systems requires overcoming unique technical challenges stemming from the diversity of use-cases, spanning from on-demand video streaming~\cite{Yeo2018} to video-conferencing~\cite{Hu2019}.





This paper provides a timely and up-to-date overview of the growing area of visual content delivery systems that employ neural enhancement. In particular, we first describe the typical architecture and major components of such systems. We then survey the state-of-the-art existing systems (Table~\ref{tab:surveyd_systems}) across diverse content delivery applications including on-demand video and image services, visual analytics, video-conferencing, live streaming and 360\textdegree~ videos. We conclude by discussing future directions for the design of neural visual content delivery systems that draw from the latest progress by the computer vision community and describe how they can be integrated to value-add existing systems.



\vspace{-0.1cm}

\section{Visual Content Delivery Systems}

\label{sec:background_vcdsys}
\begin{figure}[t]
    \centering
    \vspace{-0.25cm}
    \includegraphics[width=1\columnwidth,trim={0.5cm 11cm 11cm 0cm},clip]{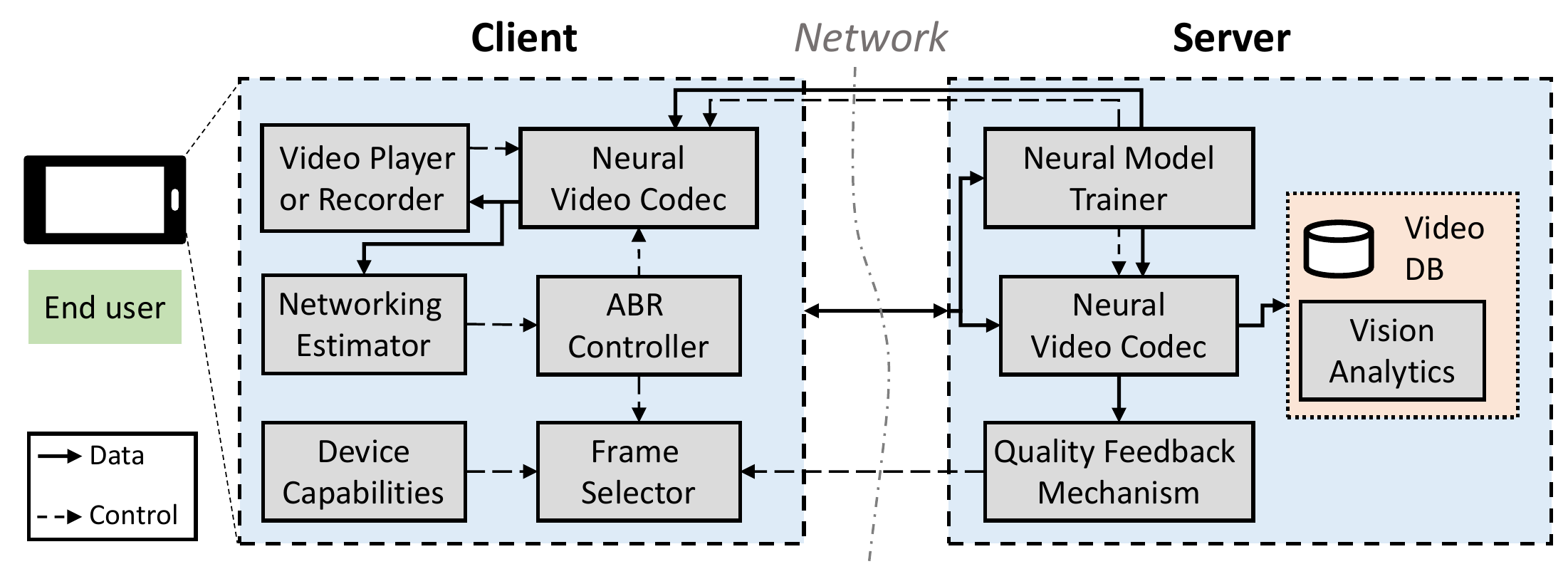}
    \vspace{-0.7cm}
    \caption{Architecture of content delivery systems.}
    \vspace{-0.5cm}
    \label{fig:content_delivery_arch}
\end{figure}

Content delivery systems are a widely studied field in systems research~\cite{distributedcache2010infocom,festive2012conext,buffer_abr2014sigcomm,Yin2015,Mao2017}. Such systems aim to deliver content to the users with minimal latency and high visual quality while supporting a diverse set of client platforms, from powerful desktops to mobile devices. Fig.~\ref{fig:content_delivery_arch} depicts the typical distributed architecture of a content delivery system. Focusing on video-on-demand without loss of generality, the client selects an online video to play and its video player app sends a request to the server side. The server fetches the video from its database, breaks it into seconds-long segments 
and begins streaming them to the client.

For such systems to meet the performance targets, the communication channel between client and server has to sustain high bandwidth throughout the streaming process. This constitutes a strong assumption that breaks for mobile clients where the connectivity conditions vary continuously. Hence, additional techniques, such as adaptive bitrate and neural enhancement, have been introduced that enable the dynamic adaptation to the varying quality of the channel.

\textbf{Adaptive Bitrate.}
To remedy the dependence of content delivery systems on network conditions, adaptive bitrate (ABR) algorithms have emerged~\cite{buffer_abr2014sigcomm,Yin2015, Mao2017}. Under this scheme, the client device first monitors its instantaneous bandwidth (\textit{Networking Estimator} in Fig.~\ref{fig:content_delivery_arch}) to assess the current network state, or the occupancy of its playback buffer. Next, the \textit{ABR Controller} tunes accordingly the per-segment bitrate that it requests from the server and configures the client's codec to decode at the selected rate. On the remote side, the server encodes each video segment with the specified bitrate and streams them to the client. Overall, ABR techniques function as a way to control the network footprint \textit{at run time} and thus helps to minimize rebuffering. 
Although ABR has been significantly improved through deep learning-based algorithms~\cite{Mao2017}, it often fails in scarce network conditions as it relies solely on network resources. 

\setlength{\tabcolsep}{2pt}
\begin{table}[t]
    \centering
    \caption{Overview of Visual Content Delivery Systems}
    \vspace{-0.3cm}
    \resizebox{\linewidth}{!}{
    \scriptsize
    \begin{tabular}{l l r c} 
        \toprule
        \begin{tabular}{@{}c@{}}\textbf{System} \\  \end{tabular} & \begin{tabular}{@{}c@{}} \textbf{Task} \\  \end{tabular} 
        & \begin{tabular}{@{}c@{}} \textbf{Ref. CNN Model} \\  \end{tabular} 
         & \begin{tabular}{@{}l@{}}  \textbf{CNN Execution}  \\  \end{tabular}
        \\ 
        \midrule
        Yeo \textit{et al.}~\cite{Yeo2017} & On-demand video delivery & VDSR~\cite{VDSR} & Client \\
        \texttt{NAS}~\cite{Yeo2018} & On-demand video delivery & MDSR~\cite{EDSR} & Client \\
        \texttt{PARSEC}~\cite{Dasari2020} & 360\textdegree~ video delivery & MDSR~\cite{EDSR} & Client \\
        \texttt{MobiSR}~\cite{Lee2019} & On-demand image delivery & RCAN~\cite{RCAN} & Client \\
        \texttt{Supremo}~\cite{Supremo2020} & On-demand image delivery & IDN~\cite{IDN} & Server \\
        \texttt{CloudSeg}~\cite{Wang2019} & Video segmentation & CARN~\cite{CARN} & Server \\
        \texttt{Dejavu}~\cite{Hu2019} & Video-conferencing & EDSR~\cite{EDSR} & Client \\
        \texttt{LiveNAS}~\cite{Kim2020} & Live streaming & MDSR~\cite{EDSR} & Server \\
        \bottomrule
    \end{tabular}
    }
    \label{tab:surveyd_systems}
    \vspace{-0.5cm}
\end{table}

\textbf{Neural Enhancement.}
%
Neural enhancement aims to restore and recover the quality/resolution of visual input. As this problem is inherently ill-posed, most works enforce a strong prior to mitigate its ill-posed nature. To this end, most state-of-the-art approaches utilize CNNs to learn the prior as it results in superior visual performance.
These methods train a model to either map a low- to a high-quality image using exemplar pairs~\cite{SRCNN,EDSR,VDSR} or exploit the internal recurring statistics of the image to enhance/upscale it~\cite{ZSSR}. 

The primary paradigm of using neural enhancement models in content delivery systems comprises the transmission of compact low-resolution/low-quality content followed by its subsequent enhancement on the receiver side through the enhance-capable models~\cite{Yeo2017, Yeo2018}. 
In this manner, the transfer load is tunable by means of the upscaling factor (for SR) and the degree of compression, controlling the system's network footprint and the associated bandwidth requirements. 
Therefore, neural enhancement opens up a new dimension in the design space by introducing a trade-off between computational and network resources, effectively overcoming existing systems' sole reliance on network resources.
To this end, existing systems may choose to independently optimize the utilization of these neural enhancement models~\cite{Lee2019,Kim2020} or integrate them within existing ABR algorithms~\cite{Yeo2018,Dasari2020}.

\textbf{Challenges of Neural Enhancement.}
Despite their advantages, neural enhancement CNNs are extremely expensive in terms of both computational and memory burden; these models are orders of magnitude larger than image discriminative models, with efficiency-optimized SR models~\cite{TPSR,FSRCNN} measured in GFLOPs, whereas their image classification counterparts~\cite{MobileNetv3} are measured in MFLOPs~\cite{embench_2019}. 
%
Although the memory footprint is hugely addressed by splitting the image into patches before processing, the computational cost is still a big challenge for real-time applications~\cite{Lee2019,Dasari2020}.

Furthermore, models trained on standard datasets that aim to generalize across all videos/images result in outputs of varying performance upon deployment and often fail catastrophically on unexpected inputs~\cite{Yeo2017}.
On the other hand, tailoring a CNN towards a specific video/image helps to mitigate this drop in performance at the cost of additional training per video/image.
In this respect, a \textit{generalization-specialization trade-off} is exposed which system designers need to decide how to control based on the target use-case.


\section{The Landscape of CNN-driven Visual Content Delivery}
\label{sec:landscape_vcdsys}

Despite their deployment barriers, several recent frameworks have incorporated neural enhancement methods into their pipelines and introduced novel techniques for overcoming their challenges. In this context, we survey the state-of-the-art visual content delivery systems that leverage neural enhancement models, taxonomizing them based on the type of content (video or image), and provide an analysis of how they counteract \circled{a}~the excessive computational requirements and \circled{b}~the performance variability across different content. 




\subsection{On-demand Content Delivery Systems}

\subsubsection{On-demand Video Streaming}

Video-on-demand (VOD) services allow users to watch content at their suitable time from any Internet-enabled device. The user selects a video to watch which is then fetched and streamed by the video server to the user device. With the majority of VOD services being interactive, on-demand video streaming systems have to yield low response time and minimal rebuffering while not compromising visual quality in order to maximize the QoE. In achieving these targets, the bottleneck lies in the link between the video server and the client with the bandwidth of the connection directly affecting the end performance.

Yeo \textit{et al.}~\cite{Yeo2017} presented one of the first works that employed neural enhancement to overcome this limitation and offered a way to utilize the clients' computational power. 
Specifically, the authors first grouped videos into clusters according to their category (basketball, athletics, \textit{etc.}) and then trained a specialized SR model, VDSR~\cite{VDSR}, for each cluster, reducing the performance variation (challenge~\circled{b}) as compared to using a single generic model. 
They also proposed the use of more compact representations such as their edges or the luminance channel, on top of tuning the spatial size of the frames, to further reduce the use of both bandwidth and computational resources.
Although these compact representations were shown not to work well in practice with the H.264 codec, they were adopted in later works, such as \texttt{Dejavu}~\cite{Hu2019}, to tackle challenge~\circled{a} in different use-cases.

To handle the excessive computational needs of neural enhancement models (challenge~\circled{a}), the authors constrained their system to work with up to 720p videos and targeted homogeneous client platforms hosting powerful desktop-grade GPUs.
This limitation was subsequently addressed to accommodate clients with heterogeneous computational capabilities in their extended proposed framework - \texttt{NAS}~\cite{Yeo2018}.

\begin{figure*}[t]
    \vspace{-0.3cm}
    \begin{subfigure}{0.245\textwidth}
        \centering
        \vspace{-0.25cm}
        \includegraphics[width=1\textwidth,trim={8cm 13.5cm 15cm 1.5cm},clip]{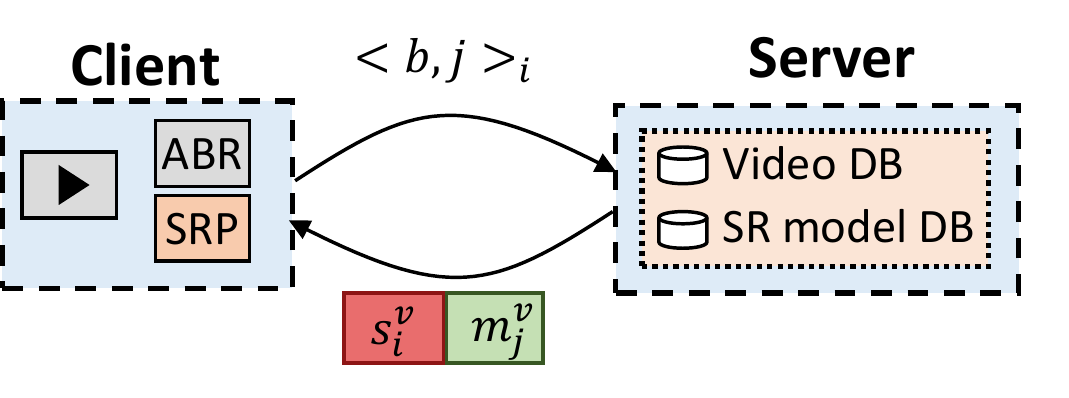}
        \caption{}
        \vspace{-0.175cm}
        \label{fig:nas_arch}
    \end{subfigure}
    \begin{subfigure}{.245\textwidth}
        \centering
        \includegraphics[width=1.3\textwidth,trim={5cm 13.5cm 11.8cm 1.5cm},clip]{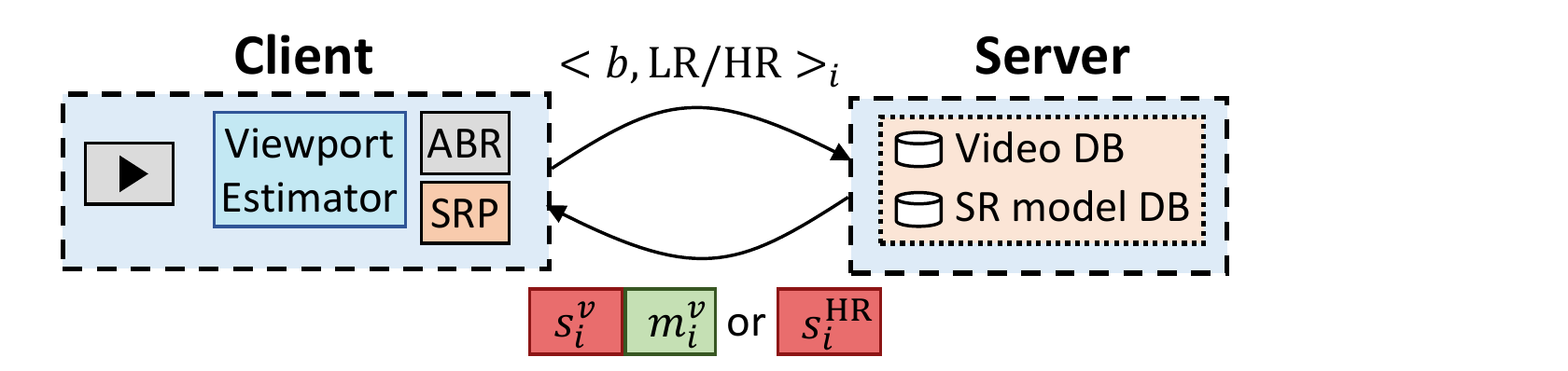}
        \caption{}
        \vspace{-0.2cm}
        \label{fig:parsec_arch}
    \end{subfigure}
    \begin{subfigure}{.245\textwidth}
        \centering
        \vspace{0.07cm}
        \includegraphics[width=1.1\textwidth,trim={6cm 13.5cm 15cm 1.8cm},clip]{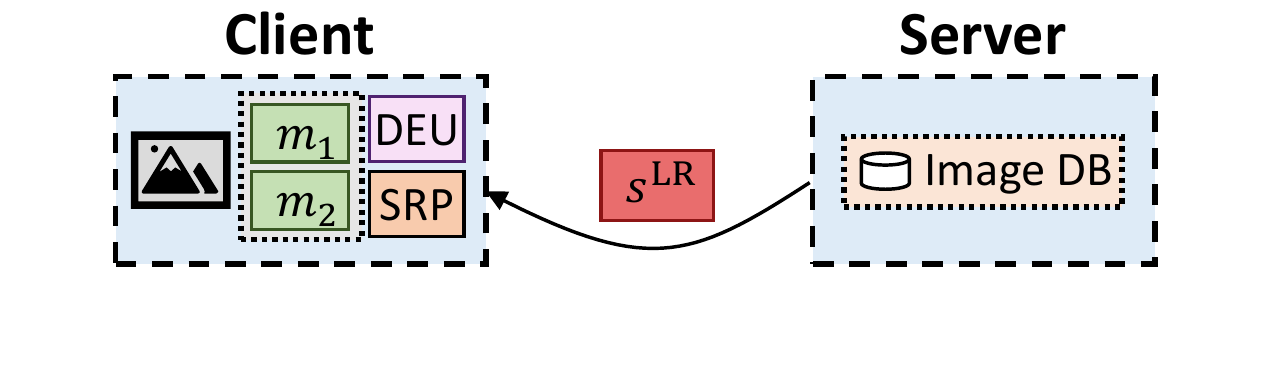}
        \vspace{-0.6cm}
        \caption{}
        \vspace{-0.2cm}
        \label{fig:mobisr_arch}
    \end{subfigure}
    \begin{subfigure}{.245\textwidth}
        \centering
        \vspace{0.07cm}
        \includegraphics[width=1.2\textwidth,trim={5cm 13.5cm 15cm 1.75cm},clip]{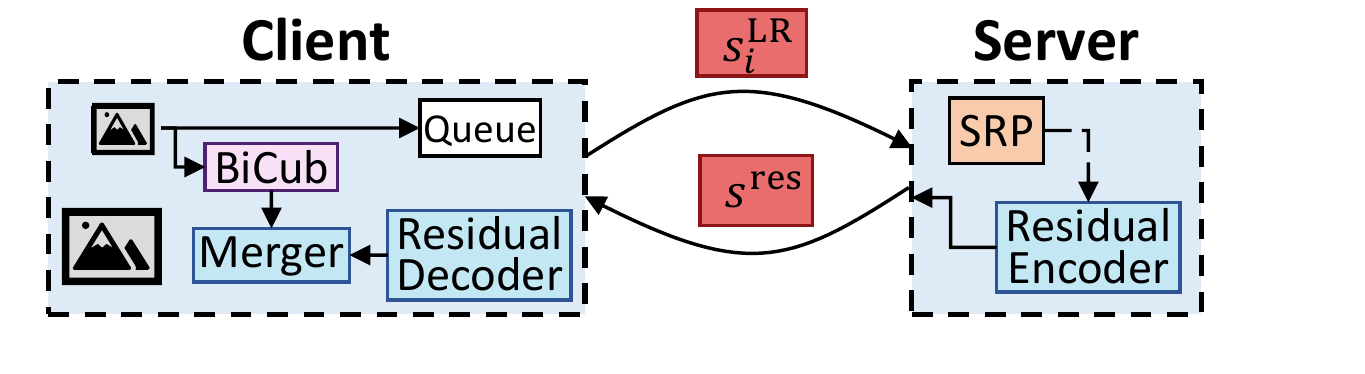}
        \vspace{-0.7cm}
        \caption{}
        \vspace{-0.2cm}
        \label{fig:supremo_arch}
    \end{subfigure}
    \vspace{-0.1cm}
    \caption{Overview of on-demand delivery systems:
    \ref{fig:nas_arch})~\texttt{NAS}~\cite{Yeo2018}, 
    \ref{fig:parsec_arch})~\texttt{PARSEC}~\cite{Dasari2020},
    \ref{fig:mobisr_arch})~\texttt{MobiSR}~\cite{Lee2019},
    \ref{fig:supremo_arch})~\texttt{Supremo}~\cite{Supremo2020}.
    }
    \vspace{-0.4cm}
\end{figure*}

In \texttt{NAS}~\cite{Yeo2018}, the authors addressed the problem of heterogeneous clients through the use of early-exit SR models of varying sizes and computational workload, allowing each client to select the appropriate model (challenge~\circled{a}) based on their resource constraints. 
To this end, they extended previous reinforcement-learning-based ABR algorithms~\cite{Mao2017} to decide not only the bitrate, but also the fraction of the SR model to be transmitted for each video segment. 
Further mitigating challenge~\circled{a}, the authors deploy a smaller variant of MDSR~\cite{EDSR} quantized at 16-bit half-precision floating-point format and execute it on a desktop-grade GPU on the client side.
Finally, instead of categorizing videos into coarse clusters as in~\cite{Yeo2017}, \texttt{NAS} tackles challenge~\circled{b} by first pre-training a generic SR model and then fine-tuning a specialized model \textit{for each video}. 
Overall, as shown in Fig.~\ref{fig:nas_arch}, the client selects the bitrate $b$ for the i-th video segment and the fraction $j$ of the SR model for the particular video and receives the i-th low-resolution segment $s_i^v$ and the associated fraction of the specialized model $m_j^v$ for video $v$ that are used by the \textit{SRP}.


\subsubsection{360\textdegree~ Video Streaming}

Compared to regular videos, streaming 360\textdegree~ videos has significantly elevated bandwidth requirements. 
To alleviate this, existing systems employ viewport prediction techniques~\cite{Fan2017} which estimate which part of the video the user will look towards and only download this spatial content. 
However, accurate viewpoint prediction is still difficult to achieve, exacerbating the problem as missing patches of the current viewpoint are needed to be fetched at the time of viewing. 
Although neural enhancement models can be utilized to mitigate these challenges, the larger spatial dimensions of 360\textdegree~ content further aggravates challenge~\circled{a} and calls for dedicated deployment solutions. 

In this context, Dasari \textit{et al.}~\cite{Dasari2020} proposed a 360\textdegree~ video streaming framework named \texttt{PARSEC} (Fig.~\ref{fig:parsec_arch}). 
Unlike previous works, the authors extended the ABR algorithm to decide on the low-resolution (LR) patches to be upscaled and the bitrate of high-resolution (HR) patches to be downloaded based on 1)~the networking conditions and the client's computational resources, 2) the viewport prediction (see \textit{Viewport Estimator} in Fig.~\ref{fig:parsec_arch}) and 3) the qualities (PSNR) of both the HR and the upsampled patches. Since the proposed ABR algorithm is designed to maximize QoE and can thus selectively decide which patches are upscaled or downloaded, challenge~\circled{b} is mitigated. To overcome challenge~\circled{a}, \texttt{PARSEC} combines an extreme upscaling factor of $\times$64 that allows ultra-LR patches to be transmitted, together with a manually tuned efficient SR model, similar to that used by \texttt{NAS}, specialized for each video segment.


\vspace{-0.4cm}
\subsubsection{On-demand Image Delivery}

As chipsets on commodity devices are gradually getting more powerful~\cite{embench_2019,ai_benchmark_2019}, this enables many applications to run fully on-device, avoiding the latency and privacy issues of offloading. 
In this direction, Lee \textit{et al.}~\cite{Lee2019} proposed \texttt{MobiSR} (Fig. \ref{fig:mobisr_arch}), a system that capitalizes over the heterogeneous compute engines of modern smartphones, \textit{e.g.} CPU, GPU and NPU, through a model selection mechanism to deliver rapid image super-resolution.

As a first step, \texttt{MobiSR} derives two model variants by applying a wide range of compression techniques on a reference model (RCAN~\cite{RCAN}) and then assigns the resulting Pareto-optimal models to the different available compute engines.
The authors showed that both large and small models perform similarly on hard-to-upscale patches, with difficulty quantified using the total-variation metric~\cite{total_variation_1992}.
Leveraging this insight, they used a hardware-aware scheduler, (\textit{Difficulty Evaluation Unit} (DEU)), to rapidly process hard-to-upscale patches using a more compact model ($m_1$) while feeding the easier patches to a larger model ($m_2$) to obtain higher quality. Hence, the image quality is maximized while meeting the applications' latency constraints (challenge~\circled{a}).

With respect to challenge~\circled{b}, \texttt{MobiSR} is optimized to achieve higher overall performance through a generic model and does not employ model specialization.
Finally, although \texttt{MobiSR} pushes on-device neural enhancement, its achieved processing rate is still not suitable for real-time video use-cases but is sufficient for reducing the data usage when using image-centric applications such as Instagram and Reddit.

\subsubsection{Cloud-assisted Image Delivery}
To accommodate real-time use-cases, Yi \textit{et al.}~\cite{Supremo2020} proposed \texttt{Supremo} (Fig.~\ref{fig:supremo_arch}), a framework that enables real-time on-device SR by selectively offloading computation to the cloud.
Following other works such as \texttt{LiveNAS}~\cite{Kim2020}, \texttt{Supremo} uses a lightweight variant of a SR model to be run on the resource-rich server, mitigating challenge~\circled{a}, and performs patch selection to only transmit key patches.
Specifically, \texttt{Supremo}'s patch selection mechanism starts by extracting the edges from each image and sorting them according to edge intensity. 
Next, depending on the networking conditions, latency requirements and their ranking, these patches are sent to the cloud to be upsampled through an SR model.
To further reduce the network footprint required to download the super-resolved patches, \texttt{Supremo} exploits the sparsity of the difference between the super-resolved patches and bicubic-upsampled patches.
As these differences are often very sparse, encoding them through the \textit{Residual Encoder} results in a heavily compressed signal, thus minimizing bandwidth.
Similar to \texttt{MobiSR}, \texttt{Supremo} handles challenge~\circled{b} by employing a generic model that aims to maximize the average upscaling performance across all processed images.

\vspace{-0.3cm}
\subsection{Live Content Streaming Systems}
\vspace{-0.1cm}

\subsubsection{Streaming for Video Analytics}
\label{sec:video_analytics}

Pipelines for video analytics~\cite{videostorm_2017,focus_2018, Du2020} perform real-time intelligent tasks over user inputs in order to enable the development of novel application such as augmented and virtual reality apps~\cite{edge_ar2019mobicom}. Such tasks span from scene labeling and object detection to face recognition. To meet the real-time performance requirements across diverse hardware platforms, such systems often rely on cloud-centric solutions. In this setup, the client device transmits the input frames to a powerful server for analysis and collects back only the result.

\begin{figure}[t]
    \begin{subfigure}{.24\textwidth}
        \centering
        \vspace{0.8cm}
        \includegraphics[width=0.85\textwidth,trim={7cm 13.5cm 15cm 1.5cm},clip]{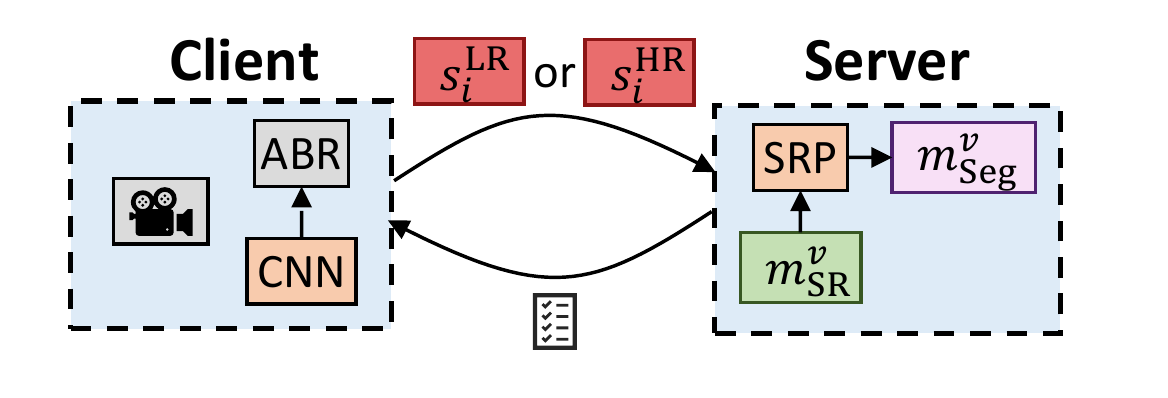}
        \caption{}
        \label{fig:cloudseg_arch}
    \end{subfigure}
    \begin{subfigure}{.23\textwidth}
        \centering
        \includegraphics[width=0.75\textwidth,trim={7.5cm 12cm 16cm 0.5cm},clip]{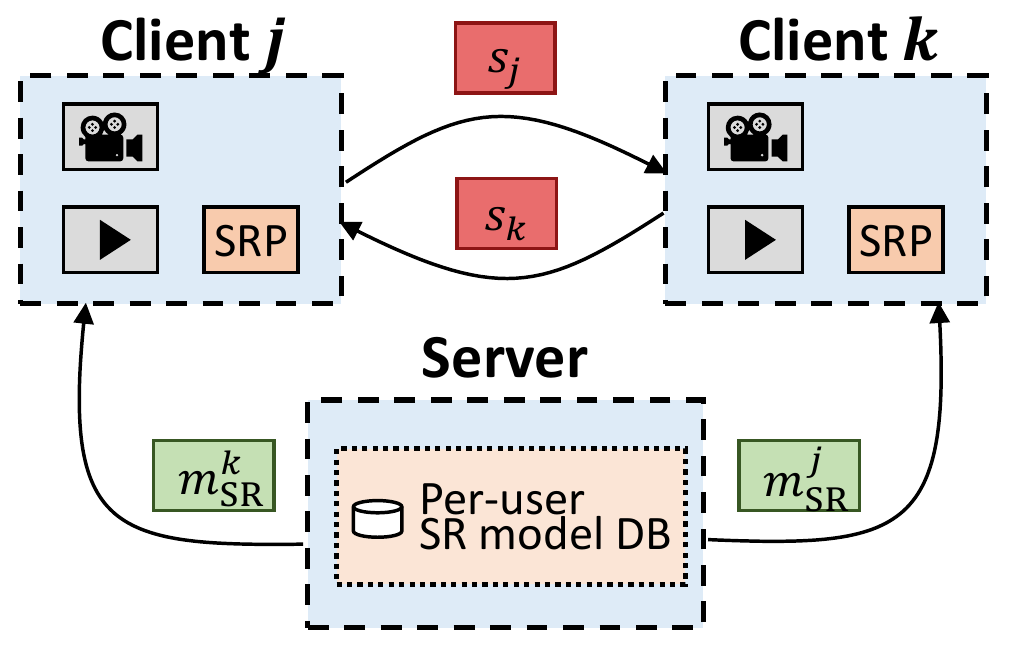}
        \caption{}
        \label{fig:dejavu_arch}
    \end{subfigure}
    \begin{subfigure}{.24\textwidth}
        \centering
        \includegraphics[width=1\textwidth,trim={5.5cm 13.5cm 15cm 1.5cm},clip]{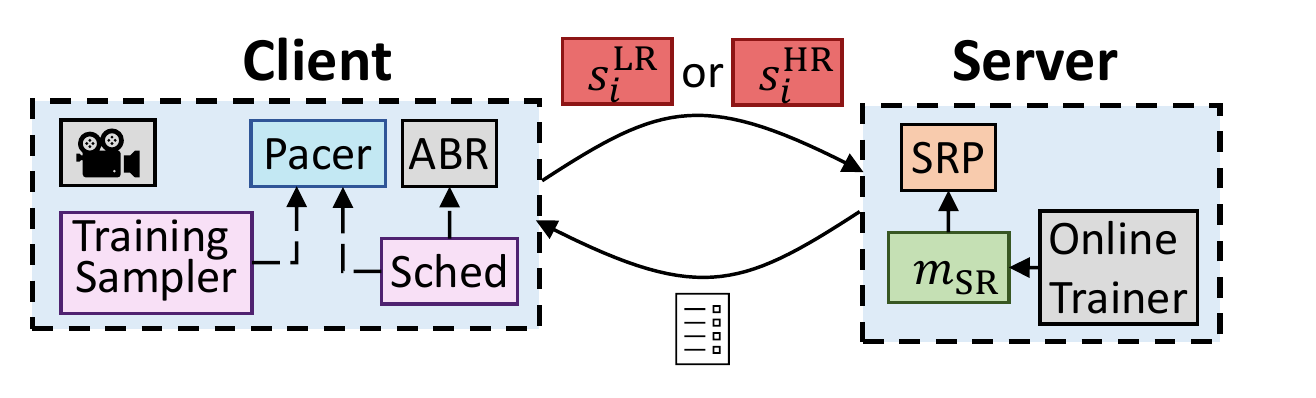}
        \caption{}
        \vspace{-0.3cm}
        \label{fig:livenas_arch}
    \end{subfigure}
    \caption{Overview of live streaming systems:
    \ref{fig:cloudseg_arch})~\texttt{CloudSeg}~\cite{Wang2019},
    \ref{fig:dejavu_arch})~\texttt{Dejavu}~\cite{Hu2019},
    \ref{fig:livenas_arch})~\texttt{LiveNAS}~\cite{Kim2020}.
    }
    \vspace{-0.5cm}
\end{figure}

Naturally, these video analytic frameworks can benefit from the transmission of lower-resolution/quality images under low-bandwidth settings. 
However, the use of low-resolution/quality images is known to reduce accuracy~\cite{Chen2015}. 
Therefore, these frameworks can use additional server-side computation and deploy neural enhancement models to minimize the accuracy loss of the targeted task. 
To achieve this goal, Wang \textit{et al.}~\cite{Wang2019} proposed \texttt{CloudSeg} (Fig.~\ref{fig:cloudseg_arch}) that jointly trains an SR model (CARN~\cite{CARN}) together with its target analytics task, \textit{i.e.} semantic segmentation (ICNet~\cite{ICNet}). 
Specifically, they updated CARN using gradients computed from the content loss between the HR and the super-resolved image and the accuracy difference between using both images in ICNet.
During inference, the SRP feeds both the LR image and super-resolved image to the pyramid segmentation model. 
Towards efficiency (challenge~\circled{a}), \texttt{CloudSeg} employs frame selection at the client side by deploying a small neural network that estimates pixel deviation in order to skip redundant stale frames. 
Finally, \texttt{CloudSeg} overcomes challenge~\circled{b} by falling back to streaming in high resolution whenever the analytics accuracy falls below a threshold.




\subsubsection{Video-Conferencing} 
To sustain the interactive communication between the callers, video-conferencing requires low response time. In an effort to achieve that, existing services~\cite{Fouladi2018} often adopt conservative strategies that relax the bandwidth requirements, but also compromise visual quality.

In this context, Hu \textit{et al.}~\cite{Hu2019} observed that in contrast to generic live streaming, video-conferencing has the property of visual similarity between \textit{recurring} sessions and designed \texttt{Dejavu} (Fig.~\ref{fig:dejavu_arch}) to exploit these unique computational and specialization opportunities.
%
The developed system starts with the \textit{offline} training of image enhancement models that are specialized \textit{per caller} (challenge~\circled{b}). In this process, the model learns to improve the video quality by increasing its \textit{encoding rate} without changing the resolution.

Upon deployment, when a video-conference session is established, the associated caller-specific enhancement model is transferred from a server to the receiver (and vice versa) ($m_{\text{SR}}^{\{k,j\}}$ in Fig.~\ref{fig:dejavu_arch}). During the call, the frames from the caller are re-encoded to lower quality prior to transmission, reducing the bandwidth usage, and the quality is enhanced on the receiver side through the caller-specific enhancement model.

To address challenge \circled{a}, \texttt{Dejavu} uses several techniques. First, a drastically scaled-down variant of the EDSR model~\cite{EDSR} is used for the quality-enhancing CNN. Next, the model is trained only on the luminance (Y) channel, instead of the typical RGB input. At the same time, a powerful GPU is assumed to be available on each calling side. Finally, \texttt{Dejavu} introduces a patch-scoring CNN that predicts the expected quality gain for each image patch. In this manner, only the top-$k$ patches that are expected to yield the highest quality improvement are processed by the quality-enhancing neural network to lower the run-time resource usage.


\subsubsection{Live Video Streaming}

In contrast to VOD services that focus on stored content, live streaming targets content produced in real-time. In this case, an additional bottleneck is introduced on the upstream client-to-server channel, as a degraded quality from the streaming user would propagate to the end users that watch the video. This property leads to additional challenges in sustaining high QoE. Moreover, while stored content in VOD or the recurrence of video-conferencing allows for offline specialization of the enhancement models, the real-time nature of live streaming requires online methods to tailor the models to the incoming video.

To tackle this problem, Kim \textit{et al.}~\cite{Kim2020} proposed the \texttt{LiveNAS} system (Fig.~\ref{fig:livenas_arch}) that focuses on optimizing the upstream transmission from streamer to server. In this system, a pre-trained generic SR model resides on the server side. Similarly to \texttt{NAS}~\cite{Yeo2018}, the selected model is a lightweight variant of MDSR~\cite{EDSR}.
Upon deployment, the streamer uploads a series of low-resolution frames which are then enhanced at the server side by the SR processor (SRP in Fig.~\ref{fig:livenas_arch}). 

To counteract the performance variability across diverse content (challenge~\circled{b}), \texttt{LiveNAS} introduces an online learning scheme that tailors the model to the particular unseen video.
This scheme consists of selectively picking high-quality patches using the \textit{Training Sampler} and transmitting them from the streamer to the server using the \textit{Pacer} (Fig.~\ref{fig:livenas_arch}). 
Since the patches needed for online training share bandwidth with the video, it is crucial to send only the patches with the highest expected impact. 
Hence, the \textit{Training Sampler} detects patches that are hard to compress with high quality, by calculating the PSNR between HR patches and their bilinearly-interpolated LR encoding and selecting the lowest-PSNR patches.
The \textit{Pacer}, on the other hand, is responsible for allocating the available bandwidth between the low-resolution patches to be upscaled and the high-resolution training patches by adaptively tuning the respective bitrate through a quality-optimizing scheduling algorithm (\textit{Sched} in Fig.~\ref{fig:livenas_arch}).

On the server side, the transferred high-resolution patches are used by the \textit{Online Trainer} to fine-tune the SR model. The invocation frequency of the Online Trainer is controlled based on an adaptive mechanism that detects training saturation or scene changes, by tracking the performance gain of the latest model as well as comparing to the initial model. Thus, the amount of training for each live stream is adapted to maximize performance without excessive resource usage.
%
%
Finally, to alleviate challenge \circled{a}, \texttt{LiveNAS} supports scale-out execution by parallelizing the SR computations across multiple GPUs (\textit{e.g.} three GPUs for 1080p to 4K real-time enhancement), if available, on the server.


\section{Future Directions}

In this section, we propose various approaches that are drawn from the latest computer vision research and provide insights on how neural enhancement can further benefit content delivery systems. 

\textbf{Visual Quality}.
One of the main open challenges in neural enhancement algorithms is the design of a metric that will correspond well with human raters. 
Distortion-based metrics, such as PSNR or SSIM~\cite{SSIM} 
have been extensively shown to improve image fidelity at the cost of perceptual quality, leading to blurry and unnatural outcomes~\cite{SRGAN}.
On the other hand, optimizing only for a perceptual-based metric such as NIQE~\cite{NIQE} and LPIPS~\cite{LPIPS}\footnote{These metrics are accessible and easily computable using code provided by the respective authors.} will lead to more natural looking images at the expense of fidelity and therefore the occasional occurrence of image artefacts. 
Mathematically, there is a trade-off between fidelity and perceptual quality~\cite{Blau_2018}.

As all the existing neural enhancement frameworks (Section~\ref{sec:landscape_vcdsys}) train their models using a distortion-based metric, the outputs of these models are accurate, but may look unnatural. 
Although this will benefit video analytics frameworks, such as \texttt{CloudSeg}, having blurry outputs will undermine the goal of other content streaming systems such as \texttt{Dejavu} and \texttt{LiveNAS}. 
To close this gap, these systems can benefit further by utilizing recent methods proposed in computer vision to train and optimize their neural enhancement models. 
These works focus on striking an optimal point between image fidelity and perceptual quality by optimizing either for both metrics~\cite{ESRGAN,TPSR}, the interpolation between a distortion-based and perceptual-based output image~\cite{ImageInterpolation} or model~\cite{ESRGAN,NetworkInterpolation}, or introducing additional priors~\cite{SPSR,NatSR} to alleviate undesirable artefacts caused by optimizing for perception-based metrics.
Finally, for video-based solutions, temporal interpolation methods~\cite{Zuckerman2020,Xiang2020} can enable a system to increase the achieved frame rate, and hence the QoE, by estimating intermediate frames rather than transferring them. 

\textbf{Efficiency-optimized Models.} 
Most neural enhancement frameworks generally adopt popular full-blown SR models (Table~\ref{tab:surveyd_systems}) and revise them in order to speed up training and inference or fit into the limiting constraints for client-side computation. 
However, these revisions are usually sub-optimal or, in some cases, detrimental. 
For instance, \texttt{PARSEC} uses of batch normalization~\cite{BN} to speed up training, reducing image fidelity~\cite{EDSR} and introducing image artefacts~\cite{ESRGAN}. 
Therefore, instead of naively scaling down and modifying full-blown SR models, these systems can leverage off-the-shelf manually-designed efficient models such as IDN~\cite{IDN} -already used by \texttt{Supremo}, automatically-designed variants such as ESRN~\cite{ESRN} and TPSR~\cite{TPSR} or even binarized SR models~\cite{Ma2019, Xin2020} in order to deliver higher-quality enhancement at a lower computational cost (challenge~\circled{a}).

\textbf{Image Rescaling.} 
One of the key benefits of deploying an SR model as a neural enhancement unit is its ability to work without the high-resolution ground-truth. 
However, in many content delivery settings, the ground-truth is available. 
Hence, many of these works can leverage the downscaling operation during the upscaling process using neural image rescaling techniques to counteract performance variability (challenge~\circled{b}) and further boost image reconstruction. 
For instance, a downscaling CNN can be trained jointly with an existing SR model as shown in~\cite{Kim2018} and techniques such as encoder-decoder frameworks and invertible neural networks can also be utilized 
as shown in \cite{Li2019} and \cite{Xiao2020} respectively. 

Despite its benefits, neural-based image rescaling incurs an additional cost of executing a downscaling neural network as compared to that of interpolation methods, utilizing additional computational resources for a more robust improvement in visual quality. 
Therefore, image rescaling techniques may be more suitable for on-demand video systems, such as \texttt{NAS} and \texttt{PARSEC}, in which the downscaling cost is an offline one-time cost across videos. 

\textbf{Meta-learning.}
To mitigate challenges~\circled{a} and \circled{b}, many systems pre-train their generic neural enhancement networks offline before fine-tuning, either offline or online, for each specific image/video.
For instance, \texttt{NAS} first trains a generic model before using its weights to fine-tune a separate model for each video in order to amortize the one-time offline training cost. 
In order to speed up and improve the performance during the fine-tuning process, these works can adopt a meta-learning approach in order to find a more optimal set of initialization parameters for fine-tuning. 
Specifically, pre-training a neural enhancement model via meta-learning on an external dataset will then require fewer gradient updates during the fine-tuning stage, therefore requiring less computational resources and leading to better performance as compared to naively fine-tuning~\cite{Soh2020,Park2020}.

\vspace{-0.2cm}
\section{Conclusion}
\label{sec:conclusion}

As the demand for content traffic grows over the upcoming years, 
the use of neural enhancement models will gain more traction in content delivery system design not only in existing applications but also emerging technologies such as augmented reality/virtual reality~\cite{overlay2015mobisys,ar2019mobicom} and telepresence~\cite{Zakharov2020}. 
Concurrently, as our everyday devices get more powerful, these models will ultimately run fully on-device, shifting the focus from accommodating stringent computational budgets to maximizing visual quality. 
By integrating ideas from both the computer vision and systems communities, we envision to align both ends towards more effective and deployable neural enhancement.

\bibliographystyle{ACM-Reference-Format}
\bibliography{references}


\begin{thebibliography}{60}


\ifx \showCODEN    \undefined \def \showCODEN     #1{\unskip}     \fi
\ifx \showDOI      \undefined \def \showDOI       #1{#1}\fi
\ifx \showISBNx    \undefined \def \showISBNx     #1{\unskip}     \fi
\ifx \showISBNxiii \undefined \def \showISBNxiii  #1{\unskip}     \fi
\ifx \showISSN     \undefined \def \showISSN      #1{\unskip}     \fi
\ifx \showLCCN     \undefined \def \showLCCN      #1{\unskip}     \fi
\ifx \shownote     \undefined \def \shownote      #1{#1}          \fi
\ifx \showarticletitle \undefined \def \showarticletitle #1{#1}   \fi
\ifx \showURL      \undefined \def \showURL       {\relax}        \fi
\providecommand\bibfield[2]{#2}
\providecommand\bibinfo[2]{#2}
\providecommand\natexlab[1]{#1}
\providecommand\showeprint[2][]{arXiv:#2}

\bibitem[\protect\citeauthoryear{Ahn, Kang, and Sohn}{Ahn
  et~al\mbox{.}}{2018}]%
        {CARN}
\bibfield{author}{\bibinfo{person}{Namhyuk Ahn}, \bibinfo{person}{Byungkon
  Kang}, {and} \bibinfo{person}{Kyung-Ah Sohn}.}
  \bibinfo{year}{2018}\natexlab{}.
\newblock \showarticletitle{{Fast, Accurate, and Lightweight Super-Resolution
  with Cascading Residual Network}}. In \bibinfo{booktitle}{\emph{European
  Conference on Computer Vision (ECCV)}}.
\newblock


\bibitem[\protect\citeauthoryear{Almeida, Laskaridis, Leontiadis, Venieris, and
  Lane}{Almeida et~al\mbox{.}}{2019}]%
        {embench_2019}
\bibfield{author}{\bibinfo{person}{Mario Almeida}, \bibinfo{person}{Stefanos
  Laskaridis}, \bibinfo{person}{Ilias Leontiadis},
  \bibinfo{person}{Stylianos~I. Venieris}, {and} \bibinfo{person}{Nicholas~D.
  Lane}.} \bibinfo{year}{2019}\natexlab{}.
\newblock \showarticletitle{{EmBench: Quantifying Performance Variations of
  Deep Neural Networks Across Modern Commodity Devices}}. In
  \bibinfo{booktitle}{\emph{The 3rd International Workshop on Deep Learning for
  Mobile Systems and Applications (EMDL)}}. 6.
\newblock


\bibitem[\protect\citeauthoryear{Blau and Michaeli}{Blau and Michaeli}{2018}]%
        {Blau_2018}
\bibfield{author}{\bibinfo{person}{Yochai Blau} {and} \bibinfo{person}{Tomer
  Michaeli}.} \bibinfo{year}{2018}\natexlab{}.
\newblock \showarticletitle{{The Perception-Distortion Tradeoff}}. In
  \bibinfo{booktitle}{\emph{IEEE/CVF Conference on Computer Vision and Pattern
  Recognition (CVPR)}}.
\newblock


\bibitem[\protect\citeauthoryear{{Borst}, {Gupta}, and {Walid}}{{Borst}
  et~al\mbox{.}}{2010}]%
        {distributedcache2010infocom}
\bibfield{author}{\bibinfo{person}{S. {Borst}}, \bibinfo{person}{V. {Gupta}},
  {and} \bibinfo{person}{A. {Walid}}.} \bibinfo{year}{2010}\natexlab{}.
\newblock \showarticletitle{{Distributed Caching Algorithms for Content
  Distribution Networks}}. In \bibinfo{booktitle}{\emph{2010 Proceedings IEEE
  INFOCOM}}.
\newblock


\bibitem[\protect\citeauthoryear{Chen, Ravindranath, Deng, Bahl, and
  Balakrishnan}{Chen et~al\mbox{.}}{2015}]%
        {Chen2015}
\bibfield{author}{\bibinfo{person}{T. Chen}, \bibinfo{person}{L. Ravindranath},
  \bibinfo{person}{Shuo Deng}, \bibinfo{person}{P. Bahl}, {and}
  \bibinfo{person}{H. Balakrishnan}.} \bibinfo{year}{2015}\natexlab{}.
\newblock \showarticletitle{Glimpse: Continuous, Real-Time Object Recognition
  on Mobile Devices}. In \bibinfo{booktitle}{\emph{SenSys '15}}.
\newblock


\bibitem[\protect\citeauthoryear{Cisco}{Cisco}{2020a}]%
        {ciscoreport}
\bibfield{author}{\bibinfo{person}{Cisco}.} \bibinfo{year}{2020}\natexlab{a}.
\newblock \bibinfo{booktitle}{\emph{{Cisco Annual Internet Report (2018–2023)
  White Paper}}}.
\newblock \bibinfo{type}{{T}echnical {R}eport}. \bibinfo{institution}{Cisco
  Systems, Inc}.
\newblock
\urldef\tempurl%
\url{https://www.cisco.com/c/en/us/solutions/collateral/executive-perspectives/annual-internet-report/white-paper-c11-741490.html}
\showURL{%
\tempurl}
\newblock
\shownote{{[Retrieved: \today]}.}


\bibitem[\protect\citeauthoryear{Cisco}{Cisco}{2020b}]%
        {ciscovni}
\bibfield{author}{\bibinfo{person}{Cisco}.} \bibinfo{year}{2020}\natexlab{b}.
\newblock \bibinfo{booktitle}{\emph{{Cisco Visual Networking Index (VNI)
  Complete Forecast Update, 2017 - 2022}}}.
\newblock \bibinfo{type}{{T}echnical {R}eport}. \bibinfo{institution}{Cisco
  Systems, Inc}.
\newblock
\urldef\tempurl%
\url{https://www.cisco.com/c/dam/m/en_us/network-intelligence/service-provider/digital-transformation/knowledge-network-webinars/pdfs/1213-business-services-ckn.pdf}
\showURL{%
\tempurl}
\newblock
\shownote{{[Retrieved: \today]}.}


\bibitem[\protect\citeauthoryear{Dasari, Bhattacharya, Vargas, Sahu,
  Balasubramanian, and Das}{Dasari et~al\mbox{.}}{2020}]%
        {Dasari2020}
\bibfield{author}{\bibinfo{person}{Mallesham Dasari}, \bibinfo{person}{A.
  Bhattacharya}, \bibinfo{person}{Santiago Vargas}, \bibinfo{person}{Pranjal
  Sahu}, \bibinfo{person}{A. Balasubramanian}, {and} \bibinfo{person}{S. Das}.}
  \bibinfo{year}{2020}\natexlab{}.
\newblock \showarticletitle{Streaming 360-Degree Videos Using
  Super-Resolution}. In \bibinfo{booktitle}{\emph{IEEE Conference on Computer
  Communications (INFOCOM)}}.
\newblock


\bibitem[\protect\citeauthoryear{Deng}{Deng}{2018}]%
        {ImageInterpolation}
\bibfield{author}{\bibinfo{person}{Xin Deng}.} \bibinfo{year}{2018}\natexlab{}.
\newblock \showarticletitle{Enhancing Image Quality via Style Transfer for
  Single Image Super-Resolution}.
\newblock \bibinfo{journal}{\emph{IEEE Signal Processing Letters}}
  (\bibinfo{year}{2018}).
\newblock


\bibitem[\protect\citeauthoryear{Dong, Loy, He, and Tang}{Dong
  et~al\mbox{.}}{2016b}]%
        {SRCNN}
\bibfield{author}{\bibinfo{person}{Chao Dong}, \bibinfo{person}{Chen~Change
  Loy}, \bibinfo{person}{Kaiming He}, {and} \bibinfo{person}{Xiaoou Tang}.}
  \bibinfo{year}{2016}\natexlab{b}.
\newblock \showarticletitle{Image Super-Resolution Using Deep Convolutional
  Networks}.
\newblock \bibinfo{journal}{\emph{IEEE Transactions on Pattern Analysis and
  Machine Intelligence (TPAMI)}} (\bibinfo{year}{2016}).
\newblock


\bibitem[\protect\citeauthoryear{Dong, Loy, and Tang}{Dong
  et~al\mbox{.}}{2016a}]%
        {FSRCNN}
\bibfield{author}{\bibinfo{person}{Chao Dong}, \bibinfo{person}{Chen~Change
  Loy}, {and} \bibinfo{person}{Xiaoou Tang}.} \bibinfo{year}{2016}\natexlab{a}.
\newblock \showarticletitle{{Accelerating the Super-Resolution Convolutional
  Neural Network}}. In \bibinfo{booktitle}{\emph{European Computer on Computer
  Vision (ECCV)}}.
\newblock


\bibitem[\protect\citeauthoryear{Du, Pervaiz, Yuan, Chowdhery, Zhang, Hoffmann,
  and Jiang}{Du et~al\mbox{.}}{2020}]%
        {Du2020}
\bibfield{author}{\bibinfo{person}{Kuntai Du}, \bibinfo{person}{Ahsan Pervaiz},
  \bibinfo{person}{Xin Yuan}, \bibinfo{person}{Aakanksha Chowdhery},
  \bibinfo{person}{Qizheng Zhang}, \bibinfo{person}{Henry Hoffmann}, {and}
  \bibinfo{person}{J. Jiang}.} \bibinfo{year}{2020}\natexlab{}.
\newblock \showarticletitle{Server-Driven Video Streaming for Deep Learning
  Inference}.
\newblock \bibinfo{journal}{\emph{SIGCOMM}} (\bibinfo{year}{2020}).
\newblock


\bibitem[\protect\citeauthoryear{Fan, Lee, Lo, Huang, Chen, and Hsu}{Fan
  et~al\mbox{.}}{2017}]%
        {Fan2017}
\bibfield{author}{\bibinfo{person}{Ching-Ling Fan}, \bibinfo{person}{J. Lee},
  \bibinfo{person}{Wen-Chih Lo}, \bibinfo{person}{C. Huang},
  \bibinfo{person}{Kuan-Ta Chen}, {and} \bibinfo{person}{C. Hsu}.}
  \bibinfo{year}{2017}\natexlab{}.
\newblock \showarticletitle{Fixation Prediction for 360 Video Streaming in
  Head-Mounted Virtual Reality}. In \bibinfo{booktitle}{\emph{Network and
  Operating System Support for Digital Audio and Video}}.
\newblock


\bibitem[\protect\citeauthoryear{Fouladi, Emmons, Orbay, Wu, Wahby, and
  Winstein}{Fouladi et~al\mbox{.}}{2018}]%
        {Fouladi2018}
\bibfield{author}{\bibinfo{person}{Sadjad Fouladi}, \bibinfo{person}{John
  Emmons}, \bibinfo{person}{Emre Orbay}, \bibinfo{person}{C. Wu},
  \bibinfo{person}{Riad~S. Wahby}, {and} \bibinfo{person}{Keith Winstein}.}
  \bibinfo{year}{2018}\natexlab{}.
\newblock \showarticletitle{Salsify: Low-Latency Network Video through Tighter
  Integration between a Video Codec and a Transport Protocol}. In
  \bibinfo{booktitle}{\emph{15th USENIX Symposium on Networked Systems Design
  and Implementation (NSDI)}}.
\newblock


\bibitem[\protect\citeauthoryear{Howard, Sandler, Chu, Chen, Chen, Tan, Wang,
  Zhu, Pang, Vasudevan, Le, and Adam}{Howard et~al\mbox{.}}{2019}]%
        {MobileNetv3}
\bibfield{author}{\bibinfo{person}{A. Howard}, \bibinfo{person}{Mark Sandler},
  \bibinfo{person}{G. Chu}, \bibinfo{person}{Liang-Chieh Chen},
  \bibinfo{person}{B. Chen}, \bibinfo{person}{M. Tan}, \bibinfo{person}{W.
  Wang}, \bibinfo{person}{Y. Zhu}, \bibinfo{person}{R. Pang},
  \bibinfo{person}{V. Vasudevan}, \bibinfo{person}{Quoc~V. Le}, {and}
  \bibinfo{person}{H. Adam}.} \bibinfo{year}{2019}\natexlab{}.
\newblock \showarticletitle{Searching for MobileNetV3}.
\newblock \bibinfo{journal}{\emph{International Conference on Computer Vision
  (ICCV)}} (\bibinfo{year}{2019}).
\newblock


\bibitem[\protect\citeauthoryear{Hsieh, Ananthanarayanan, Bodik, Venkataraman,
  Bahl, Philipose, Gibbons, and Mutlu}{Hsieh et~al\mbox{.}}{2018}]%
        {focus_2018}
\bibfield{author}{\bibinfo{person}{Kevin Hsieh}, \bibinfo{person}{Ganesh
  Ananthanarayanan}, \bibinfo{person}{Peter Bodik}, \bibinfo{person}{Shivaram
  Venkataraman}, \bibinfo{person}{Paramvir Bahl}, \bibinfo{person}{Matthai
  Philipose}, \bibinfo{person}{Phillip~B. Gibbons}, {and} \bibinfo{person}{Onur
  Mutlu}.} \bibinfo{year}{2018}\natexlab{}.
\newblock \showarticletitle{{Focus: Querying Large Video Datasets with Low
  Latency and Low Cost}}. In \bibinfo{booktitle}{\emph{Proceedings of the 12th
  USENIX Conference on Operating Systems Design and Implementation (OSDI)}}.
  \bibinfo{pages}{269--286}.
\newblock
\showISBNx{978-1-931971-47-8}


\bibitem[\protect\citeauthoryear{Hu, Misra, and Katti}{Hu
  et~al\mbox{.}}{2019}]%
        {Hu2019}
\bibfield{author}{\bibinfo{person}{Pan Hu}, \bibinfo{person}{Rakesh Misra},
  {and} \bibinfo{person}{Sachin Katti}.} \bibinfo{year}{2019}\natexlab{}.
\newblock \showarticletitle{Dejavu: Enhancing Videoconferencing with Prior
  Knowledge}. In \bibinfo{booktitle}{\emph{HotMobile}}.
\newblock


\bibitem[\protect\citeauthoryear{Huang, Johari, McKeown, Trunnell, and
  Watson}{Huang et~al\mbox{.}}{2014}]%
        {buffer_abr2014sigcomm}
\bibfield{author}{\bibinfo{person}{Te-Yuan Huang}, \bibinfo{person}{Ramesh
  Johari}, \bibinfo{person}{Nick McKeown}, \bibinfo{person}{Matthew Trunnell},
  {and} \bibinfo{person}{Mark Watson}.} \bibinfo{year}{2014}\natexlab{}.
\newblock \showarticletitle{{A Buffer-Based Approach to Rate Adaptation:
  Evidence from a Large Video Streaming Service}}. In
  \bibinfo{booktitle}{\emph{SIGCOMM}}.
\newblock


\bibitem[\protect\citeauthoryear{Hui, Wang, and Gao}{Hui et~al\mbox{.}}{2018}]%
        {IDN}
\bibfield{author}{\bibinfo{person}{Zheng Hui}, \bibinfo{person}{Xiumei Wang},
  {and} \bibinfo{person}{Xinbo Gao}.} \bibinfo{year}{2018}\natexlab{}.
\newblock \showarticletitle{Fast and Accurate Single Image Super-Resolution via
  Information Distillation Network}. In \bibinfo{booktitle}{\emph{IEEE/CVF
  Conference on Computer Vision and Pattern Recognition (CVPR)}}.
\newblock


\bibitem[\protect\citeauthoryear{Ignatov, Timofte, Kulik, Yang, Wang, Baum, Wu,
  Xu, and Van~Gool}{Ignatov et~al\mbox{.}}{2019}]%
        {ai_benchmark_2019}
\bibfield{author}{\bibinfo{person}{Andrey Ignatov}, \bibinfo{person}{Radu
  Timofte}, \bibinfo{person}{Andrei Kulik}, \bibinfo{person}{Seungsoo Yang},
  \bibinfo{person}{Ke Wang}, \bibinfo{person}{Felix Baum}, \bibinfo{person}{Max
  Wu}, \bibinfo{person}{Lirong Xu}, {and} \bibinfo{person}{Luc Van~Gool}.}
  \bibinfo{year}{2019}\natexlab{}.
\newblock \showarticletitle{AI Benchmark: All About Deep Learning on
  Smartphones in 2019}. In \bibinfo{booktitle}{\emph{International Conference
  on Computer Vision (ICCV) Workshops}}.
\newblock


\bibitem[\protect\citeauthoryear{Ioffe and Szegedy}{Ioffe and Szegedy}{2015}]%
        {BN}
\bibfield{author}{\bibinfo{person}{S. Ioffe} {and} \bibinfo{person}{Christian
  Szegedy}.} \bibinfo{year}{2015}\natexlab{}.
\newblock \showarticletitle{{Batch Normalization: Accelerating Deep Network
  Training by Reducing Internal Covariate Shift}}. In
  \bibinfo{booktitle}{\emph{International Conference on Machine Learning
  (ICML)}}.
\newblock


\bibitem[\protect\citeauthoryear{Jain, Manweiler, and Roy~Choudhury}{Jain
  et~al\mbox{.}}{2015}]%
        {overlay2015mobisys}
\bibfield{author}{\bibinfo{person}{Puneet Jain}, \bibinfo{person}{Justin
  Manweiler}, {and} \bibinfo{person}{Romit Roy~Choudhury}.}
  \bibinfo{year}{2015}\natexlab{}.
\newblock \showarticletitle{{OverLay: Practical Mobile Augmented Reality}}. In
  \bibinfo{booktitle}{\emph{Proceedings of the 13th Annual International
  Conference on Mobile Systems, Applications, and Services (MobiSys)}}.
\newblock


\bibitem[\protect\citeauthoryear{Jiang, Sekar, and Zhang}{Jiang
  et~al\mbox{.}}{2012}]%
        {festive2012conext}
\bibfield{author}{\bibinfo{person}{Junchen Jiang}, \bibinfo{person}{Vyas
  Sekar}, {and} \bibinfo{person}{Hui Zhang}.} \bibinfo{year}{2012}\natexlab{}.
\newblock \showarticletitle{{Improving Fairness, Efficiency, and Stability in
  HTTP-Based Adaptive Video Streaming with FESTIVE}}. In
  \bibinfo{booktitle}{\emph{Proceedings of the 8th International Conference on
  Emerging Networking Experiments and Technologies (CoNEXT)}}.
  \bibinfo{pages}{97–108}.
\newblock


\bibitem[\protect\citeauthoryear{Kim, Choi, Lim, and Lee}{Kim
  et~al\mbox{.}}{2018}]%
        {Kim2018}
\bibfield{author}{\bibinfo{person}{Heewon Kim}, \bibinfo{person}{Myungsub
  Choi}, \bibinfo{person}{Bee Lim}, {and} \bibinfo{person}{Kyoung~Mu Lee}.}
  \bibinfo{year}{2018}\natexlab{}.
\newblock \showarticletitle{Task-Aware Image Downscaling}. In
  \bibinfo{booktitle}{\emph{European Conference on Computer Vision (ECCV)}}.
\newblock


\bibitem[\protect\citeauthoryear{Kim, Kwon~Lee, and Mu~Lee}{Kim
  et~al\mbox{.}}{2016}]%
        {VDSR}
\bibfield{author}{\bibinfo{person}{Jiwon Kim}, \bibinfo{person}{Jung Kwon~Lee},
  {and} \bibinfo{person}{Kyoung Mu~Lee}.} \bibinfo{year}{2016}\natexlab{}.
\newblock \showarticletitle{Accurate Image Super-Resolution using Very Deep
  Convolutional Networks}. In \bibinfo{booktitle}{\emph{IEEE/CVF Conference on
  Computer Vision and Pattern Recognition (CVPR)}}.
\newblock


\bibitem[\protect\citeauthoryear{Kim, Jung, Yeo, Ye, and Han}{Kim
  et~al\mbox{.}}{2020}]%
        {Kim2020}
\bibfield{author}{\bibinfo{person}{Jae-Hong Kim}, \bibinfo{person}{Youngmok
  Jung}, \bibinfo{person}{H. Yeo}, \bibinfo{person}{Juncheol Ye}, {and}
  \bibinfo{person}{D. Han}.} \bibinfo{year}{2020}\natexlab{}.
\newblock \showarticletitle{Neural-Enhanced Live Streaming: Improving Live
  Video Ingest via Online Learning}. In \bibinfo{booktitle}{\emph{SIGCOMM}}.
\newblock


\bibitem[\protect\citeauthoryear{Ledig, Theis, Husz{\'a}r, Caballero, Aitken,
  Tejani, Totz, Wang, and Shi}{Ledig et~al\mbox{.}}{2017}]%
        {SRGAN}
\bibfield{author}{\bibinfo{person}{C. Ledig}, \bibinfo{person}{L. Theis},
  \bibinfo{person}{Ferenc Husz{\'a}r}, \bibinfo{person}{J. Caballero},
  \bibinfo{person}{Andrew Aitken}, \bibinfo{person}{Alykhan Tejani},
  \bibinfo{person}{J. Totz}, \bibinfo{person}{Zehan Wang}, {and}
  \bibinfo{person}{W. Shi}.} \bibinfo{year}{2017}\natexlab{}.
\newblock \showarticletitle{Photo-Realistic Single Image Super-Resolution Using
  a Generative Adversarial Network}. In \bibinfo{booktitle}{\emph{IEEE/CVF
  Conference on Computer Vision and Pattern Recognition (CVPR)}}.
\newblock


\bibitem[\protect\citeauthoryear{Lee, Dudziak, Abdelfattah, Venieris, Kim, Wen,
  and Lane}{Lee et~al\mbox{.}}{2020}]%
        {TPSR}
\bibfield{author}{\bibinfo{person}{Royson Lee}, \bibinfo{person}{L. Dudziak},
  \bibinfo{person}{M. Abdelfattah}, \bibinfo{person}{Stylianos~I. Venieris},
  \bibinfo{person}{H. Kim}, \bibinfo{person}{Hongkai Wen}, {and}
  \bibinfo{person}{N. Lane}.} \bibinfo{year}{2020}\natexlab{}.
\newblock \showarticletitle{Journey Towards Tiny Perceptual Super-Resolution}.
  In \bibinfo{booktitle}{\emph{European Conference on Computer Vision (ECCV)}}.
\newblock


\bibitem[\protect\citeauthoryear{Lee, Venieris, Dudziak, Bhattacharya, and
  Lane}{Lee et~al\mbox{.}}{2019}]%
        {Lee2019}
\bibfield{author}{\bibinfo{person}{Royson Lee}, \bibinfo{person}{Stylianos~I.
  Venieris}, \bibinfo{person}{L. Dudziak}, \bibinfo{person}{S. Bhattacharya},
  {and} \bibinfo{person}{N. Lane}.} \bibinfo{year}{2019}\natexlab{}.
\newblock \showarticletitle{MobiSR: Efficient On-Device Super-Resolution
  through Heterogeneous Mobile Processors}. In \bibinfo{booktitle}{\emph{The
  25th Annual International Conference on Mobile Computing and Networking
  (MobiCom)}}.
\newblock


\bibitem[\protect\citeauthoryear{Li, Liu, Li, Li, Li, and Wu}{Li
  et~al\mbox{.}}{2019}]%
        {Li2019}
\bibfield{author}{\bibinfo{person}{Y. Li}, \bibinfo{person}{D. Liu},
  \bibinfo{person}{H. Li}, \bibinfo{person}{Lianghuan Li}, \bibinfo{person}{Z.
  Li}, {and} \bibinfo{person}{F. Wu}.} \bibinfo{year}{2019}\natexlab{}.
\newblock \showarticletitle{Learning a Convolutional Neural Network for Image
  Compact-Resolution}.
\newblock \bibinfo{journal}{\emph{IEEE Transactions on Image Processing (TIP)}}
  (\bibinfo{year}{2019}).
\newblock


\bibitem[\protect\citeauthoryear{Lim, Son, Kim, Nah, and Lee}{Lim
  et~al\mbox{.}}{2017}]%
        {EDSR}
\bibfield{author}{\bibinfo{person}{Bee Lim}, \bibinfo{person}{Sanghyun Son},
  \bibinfo{person}{Heewon Kim}, \bibinfo{person}{Seungjun Nah}, {and}
  \bibinfo{person}{Kyoung~Mu Lee}.} \bibinfo{year}{2017}\natexlab{}.
\newblock \showarticletitle{{Enhanced Deep Residual Networks for Single Image
  Super-Resolution}}. In \bibinfo{booktitle}{\emph{IEEE/CVF Conference on
  Computer Vision and Pattern Recognition Workshops (CVPRW)}}.
\newblock


\bibitem[\protect\citeauthoryear{Liu, Li, and Gruteser}{Liu
  et~al\mbox{.}}{2019a}]%
        {edge_ar2019mobicom}
\bibfield{author}{\bibinfo{person}{Luyang Liu}, \bibinfo{person}{Hongyu Li},
  {and} \bibinfo{person}{Marco Gruteser}.} \bibinfo{year}{2019}\natexlab{a}.
\newblock \showarticletitle{{Edge Assisted Real-Time Object Detection for
  Mobile Augmented Reality}}. In \bibinfo{booktitle}{\emph{The 25th Annual
  International Conference on Mobile Computing and Networking (MobiCom)}}.
\newblock


\bibitem[\protect\citeauthoryear{Liu, Li, and Gruteser}{Liu
  et~al\mbox{.}}{2019b}]%
        {ar2019mobicom}
\bibfield{author}{\bibinfo{person}{Luyang Liu}, \bibinfo{person}{Hongyu Li},
  {and} \bibinfo{person}{Marco Gruteser}.} \bibinfo{year}{2019}\natexlab{b}.
\newblock \showarticletitle{{Edge Assisted Real-Time Object Detection for
  Mobile Augmented Reality}}. In \bibinfo{booktitle}{\emph{The 25th Annual
  International Conference on Mobile Computing and Networking (MobiCom)}}.
\newblock


\bibitem[\protect\citeauthoryear{Ma, Rao, Cheng, Chen, Lu, and Zhou}{Ma
  et~al\mbox{.}}{2020}]%
        {SPSR}
\bibfield{author}{\bibinfo{person}{Cheng Ma}, \bibinfo{person}{Yongming Rao},
  \bibinfo{person}{Yean Cheng}, \bibinfo{person}{Ce Chen},
  \bibinfo{person}{Jiwen Lu}, {and} \bibinfo{person}{J. Zhou}.}
  \bibinfo{year}{2020}\natexlab{}.
\newblock \showarticletitle{Structure-Preserving Super Resolution With Gradient
  Guidance}. In \bibinfo{booktitle}{\emph{IEEE/CVF Conference on Computer
  Vision and Pattern Recognition (CVPR)}}.
\newblock


\bibitem[\protect\citeauthoryear{Ma, Xiong, Hu, and Ma}{Ma
  et~al\mbox{.}}{2019}]%
        {Ma2019}
\bibfield{author}{\bibinfo{person}{Y. Ma}, \bibinfo{person}{Hongyu Xiong},
  \bibinfo{person}{Zhe Hu}, {and} \bibinfo{person}{L. Ma}.}
  \bibinfo{year}{2019}\natexlab{}.
\newblock \showarticletitle{Efficient Super Resolution Using Binarized Neural
  Network}.
\newblock \bibinfo{journal}{\emph{IEEE/CVF Conference on Computer Vision and
  Pattern Recognition Workshops (CVPRW)}} (\bibinfo{year}{2019}).
\newblock


\bibitem[\protect\citeauthoryear{Mao, Netravali, and Alizadeh}{Mao
  et~al\mbox{.}}{2017}]%
        {Mao2017}
\bibfield{author}{\bibinfo{person}{Hongzi Mao}, \bibinfo{person}{R. Netravali},
  {and} \bibinfo{person}{M. Alizadeh}.} \bibinfo{year}{2017}\natexlab{}.
\newblock \showarticletitle{Neural Adaptive Video Streaming with Pensieve}.
\newblock \bibinfo{journal}{\emph{SIGCOMM}} (\bibinfo{year}{2017}).
\newblock


\bibitem[\protect\citeauthoryear{Mittal, Soundararajan, and Bovik}{Mittal
  et~al\mbox{.}}{2013}]%
        {NIQE}
\bibfield{author}{\bibinfo{person}{Anish Mittal}, \bibinfo{person}{Rajiv
  Soundararajan}, {and} \bibinfo{person}{Alan~C. Bovik}.}
  \bibinfo{year}{2013}\natexlab{}.
\newblock \showarticletitle{{Making a ``Completely Blind" Image Quality
  Analyzer}}.
\newblock \bibinfo{journal}{\emph{IEEE Signal Processing Letters}}
  (\bibinfo{year}{2013}).
\newblock


\bibitem[\protect\citeauthoryear{Park, Yoo, Cho, Kim, and Kim}{Park
  et~al\mbox{.}}{2020}]%
        {Park2020}
\bibfield{author}{\bibinfo{person}{Seobin Park}, \bibinfo{person}{Jinsu Yoo},
  \bibinfo{person}{Donghyeon Cho}, \bibinfo{person}{Jiwon Kim}, {and}
  \bibinfo{person}{Tae~Hyun Kim}.} \bibinfo{year}{2020}\natexlab{}.
\newblock \showarticletitle{Fast Adaptation to Super-Resolution Networks via
  Meta-Learning}. In \bibinfo{booktitle}{\emph{European Conference on Computer
  Vision (ECCV)}}.
\newblock


\bibitem[\protect\citeauthoryear{Rudin, Osher, and Fatemi}{Rudin
  et~al\mbox{.}}{1992}]%
        {total_variation_1992}
\bibfield{author}{\bibinfo{person}{Leonid~I. Rudin}, \bibinfo{person}{Stanley
  Osher}, {and} \bibinfo{person}{Emad Fatemi}.}
  \bibinfo{year}{1992}\natexlab{}.
\newblock \showarticletitle{{Nonlinear Total Variation Based Noise Removal
  Algorithms}}.
\newblock \bibinfo{journal}{\emph{Phys. D}} \bibinfo{volume}{60},
  \bibinfo{number}{1-4} (\bibinfo{date}{Nov.} \bibinfo{year}{1992}),
  \bibinfo{pages}{259--268}.
\newblock
\showISSN{0167-2789}


\bibitem[\protect\citeauthoryear{Shocher, Cohen, and Irani}{Shocher
  et~al\mbox{.}}{2018}]%
        {ZSSR}
\bibfield{author}{\bibinfo{person}{Assaf Shocher}, \bibinfo{person}{N. Cohen},
  {and} \bibinfo{person}{M. Irani}.} \bibinfo{year}{2018}\natexlab{}.
\newblock \showarticletitle{"Zero-Shot" Super-Resolution Using Deep Internal
  Learning}. In \bibinfo{booktitle}{\emph{IEEE/CVF Conference on Computer
  Vision and Pattern Recognition (CVPR)}}.
\newblock


\bibitem[\protect\citeauthoryear{Soh, Cho, and Cho}{Soh et~al\mbox{.}}{2020}]%
        {Soh2020}
\bibfield{author}{\bibinfo{person}{Jae~Woong Soh}, \bibinfo{person}{Sunwoo
  Cho}, {and} \bibinfo{person}{N.~I. Cho}.} \bibinfo{year}{2020}\natexlab{}.
\newblock \showarticletitle{Meta-Transfer Learning for Zero-Shot
  Super-Resolution}.
\newblock \bibinfo{journal}{\emph{IEEE/CVF Conference on Computer Vision and
  Pattern Recognition (CVPR)}}.
\newblock


\bibitem[\protect\citeauthoryear{Soh, Park, Jo, and Cho}{Soh
  et~al\mbox{.}}{2019}]%
        {NatSR}
\bibfield{author}{\bibinfo{person}{Jae~Woong Soh}, \bibinfo{person}{G.~Y.
  Park}, \bibinfo{person}{Junho Jo}, {and} \bibinfo{person}{N.~I. Cho}.}
  \bibinfo{year}{2019}\natexlab{}.
\newblock \showarticletitle{Natural and Realistic Single Image Super-Resolution
  With Explicit Natural Manifold Discrimination}. In
  \bibinfo{booktitle}{\emph{IEEE/CVF Conference on Computer Vision and Pattern
  Recognition (CVPR)}}.
\newblock


\bibitem[\protect\citeauthoryear{Song, Xu, Jia, Chen, Xu, and Wang}{Song
  et~al\mbox{.}}{2020}]%
        {ESRN}
\bibfield{author}{\bibinfo{person}{Dehua Song}, \bibinfo{person}{Chang Xu},
  \bibinfo{person}{Xu Jia}, \bibinfo{person}{Yiyi Chen},
  \bibinfo{person}{Chunjing Xu}, {and} \bibinfo{person}{Yunhe Wang}.}
  \bibinfo{year}{2020}\natexlab{}.
\newblock \showarticletitle{Efficient Residual Dense Block Search for Image
  Super-Resolution}. In \bibinfo{booktitle}{\emph{Association for the
  Advancement of Artificial Intelligence (AAAI)}}.
\newblock


\bibitem[\protect\citeauthoryear{Wang, Yu, Dong, Tang, and Loy}{Wang
  et~al\mbox{.}}{2019b}]%
        {NetworkInterpolation}
\bibfield{author}{\bibinfo{person}{Xintao Wang}, \bibinfo{person}{K. Yu},
  \bibinfo{person}{C. Dong}, \bibinfo{person}{X. Tang}, {and}
  \bibinfo{person}{Chen~Change Loy}.} \bibinfo{year}{2019}\natexlab{b}.
\newblock \showarticletitle{Deep Network Interpolation for Continuous Imagery
  Effect Transition}. In \bibinfo{booktitle}{\emph{IEEE/CVF Conference on
  Computer Vision and Pattern Recognition (CVPR)}}.
\newblock


\bibitem[\protect\citeauthoryear{Wang, Yu, Wu, Gu, Liu, Dong, Loy, Qiao, and
  Tang}{Wang et~al\mbox{.}}{2018}]%
        {ESRGAN}
\bibfield{author}{\bibinfo{person}{Xintao Wang}, \bibinfo{person}{Ke Yu},
  \bibinfo{person}{Shixiang Wu}, \bibinfo{person}{Jinjin Gu},
  \bibinfo{person}{Yihao Liu}, \bibinfo{person}{Chao Dong},
  \bibinfo{person}{Chen~Change Loy}, \bibinfo{person}{Yu Qiao}, {and}
  \bibinfo{person}{Xiaoou Tang}.} \bibinfo{year}{2018}\natexlab{}.
\newblock \showarticletitle{{ESRGAN: Enhanced Super-Resolution Generative
  Adversarial Networks}}. In \bibinfo{booktitle}{\emph{European Conference on
  Computer Vision Workshops (ECCVW)}}.
\newblock


\bibitem[\protect\citeauthoryear{Wang, Wang, Zhang, Jiang, and Chen}{Wang
  et~al\mbox{.}}{2019a}]%
        {Wang2019}
\bibfield{author}{\bibinfo{person}{Yiding Wang}, \bibinfo{person}{Weiyan Wang},
  \bibinfo{person}{Junxue Zhang}, \bibinfo{person}{J. Jiang}, {and}
  \bibinfo{person}{K. Chen}.} \bibinfo{year}{2019}\natexlab{a}.
\newblock \showarticletitle{Bridging the Edge-Cloud Barrier for Real-time
  Advanced Vision Analytics}. In \bibinfo{booktitle}{\emph{HotCloud}}.
\newblock


\bibitem[\protect\citeauthoryear{Wang, Bovik, Sheikh, and Simoncelli}{Wang
  et~al\mbox{.}}{2004}]%
        {SSIM}
\bibfield{author}{\bibinfo{person}{Zhou Wang}, \bibinfo{person}{Alan~C. Bovik},
  \bibinfo{person}{Hamid~R. Sheikh}, {and} \bibinfo{person}{Eero~P.
  Simoncelli}.} \bibinfo{year}{2004}\natexlab{}.
\newblock \showarticletitle{{Image quality assessment: from error visibility to
  structural similarity}}.
\newblock \bibinfo{journal}{\emph{IEEE Transactions on Image Processing (TIP)}}
  (\bibinfo{year}{2004}).
\newblock


\bibitem[\protect\citeauthoryear{Xiang, Tian, Zhang, Fu, Allebach, and
  Xu}{Xiang et~al\mbox{.}}{2020}]%
        {Xiang2020}
\bibfield{author}{\bibinfo{person}{X. Xiang}, \bibinfo{person}{Yapeng Tian},
  \bibinfo{person}{Yulun Zhang}, \bibinfo{person}{Y. Fu}, \bibinfo{person}{J.
  Allebach}, {and} \bibinfo{person}{Chenliang Xu}.}
  \bibinfo{year}{2020}\natexlab{}.
\newblock \showarticletitle{Zooming Slow-Mo: Fast and Accurate One-Stage
  Space-Time Video Super-Resolution}.
\newblock \bibinfo{journal}{\emph{IEEE/CVF Conference on Computer Vision and
  Pattern Recognition (CVPR)}} (\bibinfo{year}{2020}).
\newblock


\bibitem[\protect\citeauthoryear{Xiao, Zheng, Liu, Wang, He, Ke, Bian, Lin, and
  Liu}{Xiao et~al\mbox{.}}{2020}]%
        {Xiao2020}
\bibfield{author}{\bibinfo{person}{Mingqing Xiao}, \bibinfo{person}{Shuxin
  Zheng}, \bibinfo{person}{Chang Liu}, \bibinfo{person}{Yaolong Wang},
  \bibinfo{person}{Di He}, \bibinfo{person}{Guolin Ke}, \bibinfo{person}{Jiang
  Bian}, \bibinfo{person}{Zhouchen Lin}, {and} \bibinfo{person}{Tie-Yan Liu}.}
  \bibinfo{year}{2020}\natexlab{}.
\newblock \showarticletitle{Invertible Image Rescaling}. In
  \bibinfo{booktitle}{\emph{European Conference on Computer Vision (ECCV)}}.
\newblock


\bibitem[\protect\citeauthoryear{Xin, Wang, Jiang, Li, Huang, and Gao}{Xin
  et~al\mbox{.}}{2020}]%
        {Xin2020}
\bibfield{author}{\bibinfo{person}{Jingwei Xin}, \bibinfo{person}{Nannan Wang},
  \bibinfo{person}{Xinrui Jiang}, \bibinfo{person}{Jie Li},
  \bibinfo{person}{Heng Huang}, {and} \bibinfo{person}{Xinbo Gao}.}
  \bibinfo{year}{2020}\natexlab{}.
\newblock \showarticletitle{Binarized Neural Network for Single Image Super
  Resolution}. In \bibinfo{booktitle}{\emph{European Conference on Computer
  Vision (ECCV)}}.
\newblock


\bibitem[\protect\citeauthoryear{Yeo, Do, and Han}{Yeo et~al\mbox{.}}{2017}]%
        {Yeo2017}
\bibfield{author}{\bibinfo{person}{H. Yeo}, \bibinfo{person}{Sunghyun Do},
  {and} \bibinfo{person}{D. Han}.} \bibinfo{year}{2017}\natexlab{}.
\newblock \showarticletitle{How will Deep Learning Change Internet Video
  Delivery?}. In \bibinfo{booktitle}{\emph{HotNets}}.
\newblock


\bibitem[\protect\citeauthoryear{Yeo, Jung, Kim, Shin, and Han}{Yeo
  et~al\mbox{.}}{2018}]%
        {Yeo2018}
\bibfield{author}{\bibinfo{person}{H. Yeo}, \bibinfo{person}{Youngmok Jung},
  \bibinfo{person}{Jaehong Kim}, \bibinfo{person}{Jinwoo Shin}, {and}
  \bibinfo{person}{D. Han}.} \bibinfo{year}{2018}\natexlab{}.
\newblock \showarticletitle{Neural Adaptive Content-aware Internet Video
  Delivery}. In \bibinfo{booktitle}{\emph{13th {USENIX} Symposium on Operating
  Systems Design and Implementation ({OSDI})}}.
\newblock


\bibitem[\protect\citeauthoryear{Yi, Kim, Kim, and Choi}{Yi
  et~al\mbox{.}}{2020}]%
        {Supremo2020}
\bibfield{author}{\bibinfo{person}{Junheon Yi}, \bibinfo{person}{Seongwon Kim},
  \bibinfo{person}{Joongheon Kim}, {and} \bibinfo{person}{Sunghyun Choi}.}
  \bibinfo{year}{2020}\natexlab{}.
\newblock \showarticletitle{{Supremo: Cloud-Assisted Low-Latency
  Super-Resolution in Mobile Devices}}.
\newblock \bibinfo{journal}{\emph{IEEE Transactions on Mobile Computing (TMC)}}
  (\bibinfo{year}{2020}).
\newblock


\bibitem[\protect\citeauthoryear{Yin, Jindal, Sekar, and Sinopoli}{Yin
  et~al\mbox{.}}{2015}]%
        {Yin2015}
\bibfield{author}{\bibinfo{person}{Xiaoqi Yin}, \bibinfo{person}{A. Jindal},
  \bibinfo{person}{V. Sekar}, {and} \bibinfo{person}{B. Sinopoli}.}
  \bibinfo{year}{2015}\natexlab{}.
\newblock \showarticletitle{A Control-Theoretic Approach for Dynamic Adaptive
  Video Streaming over HTTP}. In \bibinfo{booktitle}{\emph{SIGCOMM}}.
\newblock


\bibitem[\protect\citeauthoryear{Zakharov, Ivakhnenko, Shysheya, and
  Lempitsky}{Zakharov et~al\mbox{.}}{2020}]%
        {Zakharov2020}
\bibfield{author}{\bibinfo{person}{E. Zakharov}, \bibinfo{person}{Aleksei
  Ivakhnenko}, \bibinfo{person}{Aliaksandra Shysheya}, {and}
  \bibinfo{person}{V. Lempitsky}.} \bibinfo{year}{2020}\natexlab{}.
\newblock \showarticletitle{{Fast Bi-layer Neural Synthesis of One-Shot
  Realistic Head Avatars}}. In \bibinfo{booktitle}{\emph{European Conference on
  Computer Vision (ECCV)}}.
\newblock


\bibitem[\protect\citeauthoryear{Zhang, Ananthanarayanan, Bodik, Philipose,
  Bahl, and Freedman}{Zhang et~al\mbox{.}}{2017}]%
        {videostorm_2017}
\bibfield{author}{\bibinfo{person}{Haoyu Zhang}, \bibinfo{person}{Ganesh
  Ananthanarayanan}, \bibinfo{person}{Peter Bodik}, \bibinfo{person}{Matthai
  Philipose}, \bibinfo{person}{Paramvir Bahl}, {and}
  \bibinfo{person}{Michael~J. Freedman}.} \bibinfo{year}{2017}\natexlab{}.
\newblock \showarticletitle{{Live Video Analytics at Scale with Approximation
  and Delay-tolerance}}. In \bibinfo{booktitle}{\emph{Proceedings of the 14th
  USENIX Conference on Networked Systems Design and Implementation (NSDI)}}.
  \bibinfo{pages}{377--392}.
\newblock
\showISBNx{978-1-931971-37-9}


\bibitem[\protect\citeauthoryear{Zhang, Isola, Efros, Shechtman, and
  Wang}{Zhang et~al\mbox{.}}{2018a}]%
        {LPIPS}
\bibfield{author}{\bibinfo{person}{Richard Zhang}, \bibinfo{person}{Phillip
  Isola}, \bibinfo{person}{Alexei~A Efros}, \bibinfo{person}{Eli Shechtman},
  {and} \bibinfo{person}{Oliver Wang}.} \bibinfo{year}{2018}\natexlab{a}.
\newblock \showarticletitle{{The Unreasonable Effectiveness of Deep Features as
  a Perceptual Metric}}. In \bibinfo{booktitle}{\emph{IEEE/CVF Conference on
  Computer Vision and Pattern Recognition (CVPR)}}.
\newblock


\bibitem[\protect\citeauthoryear{Zhang, Li, Li, Wang, Zhong, and Fu}{Zhang
  et~al\mbox{.}}{2018b}]%
        {RCAN}
\bibfield{author}{\bibinfo{person}{Yulun Zhang}, \bibinfo{person}{Kunpeng Li},
  \bibinfo{person}{Kai Li}, \bibinfo{person}{Lichen Wang},
  \bibinfo{person}{Bineng Zhong}, {and} \bibinfo{person}{Yun Fu}.}
  \bibinfo{year}{2018}\natexlab{b}.
\newblock \showarticletitle{{Image Super-Resolution Using Very Deep Residual
  Channel Attention Networks}}. In \bibinfo{booktitle}{\emph{European
  Conference on Computer Vision (ECCV)}}.
\newblock


\bibitem[\protect\citeauthoryear{Zhao, Qi, Shen, Shi, and Jia}{Zhao
  et~al\mbox{.}}{2018}]%
        {ICNet}
\bibfield{author}{\bibinfo{person}{Hengshuang Zhao}, \bibinfo{person}{Xiaojuan
  Qi}, \bibinfo{person}{Xiaoyong Shen}, \bibinfo{person}{J. Shi}, {and}
  \bibinfo{person}{J. Jia}.} \bibinfo{year}{2018}\natexlab{}.
\newblock \showarticletitle{ICNet for Real-Time Semantic Segmentation on
  High-Resolution Images}. In \bibinfo{booktitle}{\emph{European Conference on
  Computer Vision (ECCV)}}.
\newblock


\bibitem[\protect\citeauthoryear{Zuckerman, Bagon, Naor, Pisha, and
  Irani}{Zuckerman et~al\mbox{.}}{2020}]%
        {Zuckerman2020}
\bibfield{author}{\bibinfo{person}{Liad~Pollak Zuckerman}, \bibinfo{person}{S.
  Bagon}, \bibinfo{person}{Eyal Naor}, \bibinfo{person}{George Pisha}, {and}
  \bibinfo{person}{M. Irani}.} \bibinfo{year}{2020}\natexlab{}.
\newblock \showarticletitle{Across Scales \& Across Dimensions: Temporal
  Super-Resolution using Deep Internal Learning}. In
  \bibinfo{booktitle}{\emph{European Conference on Computer Vision (ECCV)}}.
\newblock


\end{thebibliography}

\end{document}